\documentclass[journal,transmag]{IEEEtran}
\usepackage{multirow}
\usepackage{graphicx}

\usepackage{amsmath}

\ifCLASSOPTIONcompsoc
 \usepackage[caption=false,font=normalsize,labelfont=sf,textfont=sf]{subfig}
\else
 \usepackage[caption=false,font=footnotesize]{subfig}
\fi

\begin{document}

\title{KST-GCN: A Knowledge-Driven Spatial-Temporal Graph Convolutional Network for Traffic Forecasting}


\author{Jiawei Zhu, Xing Han, Hanhan Deng, Chao Tao, Ling Zhao,\\ Pu Wang,~\IEEEmembership{Member,~IEEE,} Tao Lin, Haifeng Li*,~\IEEEmembership{Member,~IEEE,}
\IEEEcompsocitemizethanks{\IEEEcompsocthanksitem J. Zhu, C. Tao, X. Han, L. Zhao and H. Li are with School of Geosciences and Info-Physics, Central South University, Changsha 410083, China.
\IEEEcompsocthanksitem H. Deng  is with Hisense TransTech Co., Ltd., No.17 Donghai West Road, Qingdao, China
\IEEEcompsocthanksitem P. Wang is with with School of Traffic Transportation Engineering, Central South University, Changsha 410083, China.
\IEEEcompsocthanksitem T. Lin is with College of Biosystems Engineering and Food Science, Zhejiang University, Hangzhou, China.}
\thanks{Corresponding author: Haifeng Li, Email:lihaifeng@csu.edu.cn}

\thanks{Citation: Zhu, J.; Han, X.; Deng, H.; Tao, C.; Zhao, L.; Lin, T.; Li, H.. KST-GCN: A Knowledge-Driven Spatial-Temporal Graph Convolutional Network for Traffic Forecasting. IEEE Transactions on Intelligent Transportation Systems. 2021. doi:10.109/TITS.2021.3136287.}

}

\markboth{}%
{}

\IEEEtitleabstractindextext{%
\begin{abstract}
While considering the spatial and temporal features of traffic, capturing the impacts of various external factors on travel is an essential step towards achieving accurate traffic forecasting. However, existing studies seldom consider external factors or neglect the effect of the complex correlations among external factors on traffic. Intuitively, knowledge graphs can naturally describe these correlations. Since knowledge graphs and traffic networks are essentially heterogeneous networks, it is challenging to integrate the information in both networks. On this background, this study presents a knowledge representation-driven traffic forecasting method based on spatial-temporal graph convolutional networks. We first construct a knowledge graph for traffic forecasting and derive knowledge representations by a knowledge representation learning method named KR-EAR. Then, we propose the Knowledge Fusion Cell (KF-Cell) to combine the knowledge and traffic features as the input of a spatial-temporal graph convolutional backbone network. Experimental results on the real-world dataset show that our strategy enhances the forecasting performances of backbones at various prediction horizons. The ablation and perturbation analysis further verify the effectiveness and robustness of the proposed method. To the best of our knowledge, this is the first study that constructs and utilizes a knowledge graph to facilitate traffic forecasting; it also offers a promising direction to integrate external information and spatial-temporal information for traffic forecasting. The source code is available at https://github.com/lehaifeng/T-GCN/tree/master/KST-GCN.
\end{abstract}

\begin{IEEEkeywords}
Traffic flow forecasting, multisource spatial-temporal data, knowledge representation, spatial-temporal graph convolutional networks.
\end{IEEEkeywords}}

\maketitle

\IEEEdisplaynontitleabstractindextext

\IEEEpeerreviewmaketitle

\section{Introduction}

\IEEEPARstart{W}{ith} the steady increase in vehicle ownership, transportation demands also gradually increase. As a result, a series of problems such as traffic congestion and traffic accidents become significant. The emergence of intelligent transportation systems (ITSs) can effectively solve these problems \cite{zhang2011data}. As one of the critical technologies intelligent transportation systems, traffic flow forecasting has become a popular research topic. First, it provides data support and suggestions for urban management. Second, it also provides travelers with reliable traffic prediction reports to develop optimal routes, saving travelers time and improving travel efficiency.

Traffic forecasting is to predict the traffic flow states over a future period of time based on historical traffic information. Traffic flow states have a substantial spatial and temporal correlation \cite{Nagy2018survey,barros2015short}, which is affected not only by the previous traffic conditions of the upstream and the downstream traffic flows of the current monitoring point but also by the historical traffic conditions of the adjacent roads. With the development of deep learning, a large number of researchers have used recurrent neural networks, such as the long short-term memory (LSTM) network \cite{Ma2015long, Tian2015predicting} and the gated recurrent unit (GRU) \cite{fu2016using} network, to model the temporal dependence of traffic flows. To characterize the spatial dependence of traffic flows, some studies use convolutional neural networks (CNNs) to extract spatial information and combine them with LSTM \cite{Liu2017short} to improve the prediction accuracy. In recent years, GNNs \cite{zhou2018graph}, which are applicable to non-Euclidean structures such as road networks, have emerged to better model the spatial dependence of roads and improve the prediction accuracy \cite{zhao2019t}.

In addition, traffic information may be affected by multiple external factors, such as weather conditions, the presence of transportation stations, emergency events, holidays, and the distribution of nearby POIs \cite{Lana2018road}. These external factors have either direct or indirect relations with the traffic information that can influence the traffic conditions in the city. However, the few existing studies that consider external factors \cite{liao2018deep,zhang2018combining} simply consider external factors and ignore the influence of the interrelationships between traffic information and external factors on traffic. For example, the weather changes over time, and the traffic flows in different weather conditions may have different states. Nonetheless, road sections are not uniformly affected by the weather, and we should consider other attributes of road sections. For example, the less popular road sections with fewer surrounding facilities have less road load, so they are less affected by heavy rain than popular road sections downtown. How to integrate the semantic correlation of multisource data is the key to improving the ability to predict traffic flows. 

In recent years, the emergence of knowledge graphs has provided broader ideas for the above problem. Therefore, we first represent traffic information and multiple external factors as a heterogeneous semantic network, that is, a knowledge graph. Then, we adopt knowledge graph embedding methods to capture the semantic relationships between traffic information and external factors. Finally, we propose the Knowledge Fusion Cell (KF-Cell) to introduce the derived knowledge into backbone spatial-temporal graph convolutional networks. Our contributions are as follows:

\begin{enumerate}
    \item In order to consider the semantic correlations among various external factors and traffic information, we specially design a knowledge graph for traffic forecasting. Based on that, we design the KF-Cell to incorporate knowledge derived by the knowledge graph embedding method in spatiotemporal graph convolutional networks. 
    \item We evaluate the proposed method on a real-world dataset. Under different prediction horizons, incorporating KF-Cell enhances the traffic forecasting performances of various backbone models.
    \item We conduct ablation experiments to further prove the validity of semantic relationships in traffic forecasting and verify the influence of dynamic and static external factors on traffic forecasting.
\end{enumerate}

\section{Related Work}
\subsection{Traffic Flow Forecasting}
Conventional prediction models, including historical averages, time series, and Kalman filtering, often use statistical analysis to predict traffic conditions. The historical average model directly utilizes the average value of historical data as the prediction result. The time series model uses the relationship between current data and historical data and considers the periodicity and the tendency of the data to make predictions. The ARMA model \cite{ahmed1979analysis} proposed in 1979 is an important method to study time series. It consists of an autoregressive (AR) model and a moving average (MA) model. The AR model uses the autocorrelation function to find model parameters and predict the time series using original historical data, while the MA model accumulates the error term of the autocorrelation function. The ARIMA model \cite{hamed1995short} is a generalized version of the ARMA with an additional component of automatic differentiation, and both the ARIMA and ARMA models take the stationarity of the time series as a starting point. The Kalman filtering model uses a state space defined by a state equation and an observation to filter out the noise to make predictions.

With the continuous development of machine learning and deep learning, the advantages of intelligent forecasting models are becoming increasingly more prominent. These models take a large quantity of collected historical traffic data as the input and automatically learn the potential patterns and features in the data to predict traffic states. Intelligent forecasting models can be mainly divided into two categories: conventional machine learning approaches and deep learning approaches. As one of the most used approaches, neural networks learn the nonlinear relations in the input data to make predictions. Artificial neural networks (ANNs) and support vector regression (SVR) are two common models for practical prediction tasks. The SVR learns nonlinear statistical patterns using sufficient features from historical data. The k-nearest neighbors and fuzzy logic models are two additional examples of nonlinear parametric models. Alternatively, an ANN \cite{jun2008research} adjusts its weights and biases via backpropagation or a radial basis function (RBF) \cite{kuang2004short} and obtains linear prediction results after applying a nonlinear activation function.

The models introduced above use historical traffic state data to predict the future. For a road network composed of many road section nodes, the adjacency between road section nodes will either directly or indirectly affect the final prediction. The Bayesian network (BN) analyzes the adjacency relationships in road networks to predict traffic conditions. Another model capable of using topological information from road networks is the graph convolutional network (GCN), whose input consists of an adjacency matrix and a feature matrix. The adjacency matrix provides the topological features of a road network, and the feature matrix includes traffic information. The GCN captures the connection relationships between road section nodes to forecast future traffic conditions. However, these models only retain information about the spatial relationships in road networks and lack the capability to capture the temporal relationships. Correspondingly, models such as the feed-forward NN \cite{park1999forecasting}, the DBN \cite{huang2014deep}, the RNN \cite{zhene2018deep} and the RNN variants GRU \cite{fu2016using}, and LSTM \cite{van2002freeway} capture the tendencies and periodicity of traffic features, but they ignore the intrinsic topological characteristics of the urban traffic network. Many researchers have noticed this issue, and numerous spatial-temporal forecasting models that fully utilize both the topological structures of networks and the temporal dependence in traffic data have been proposed. Such models include the ST-ResNet \cite{zhang2016deep}, SAE \cite{lv2014traffic}, FCL-Net \cite{yu2017spatio}, DCRNN \cite{yu2017spatiotemporal} and T-GCN \cite{zhao2019t}, among others.

In addition to historical traffic information, traffic states may be affected by a variety of external factors, such as weather conditions, metro station, and bus stop information, POIs, and other factors. The main challenge of the current traffic forecasting task is to integrate external factor information into prediction models. Some methods that consider multisource data have been proposed in previous studies. Liao B et al. \cite{liao2018deep} encoded external information by using an encoder based on LSTM \cite{van2002freeway} and treated the integrated multimodal data as the input sequence of the prediction model. Based on GRU, the model proposed by Da Z et al. \cite{zhang2018combining} fuses the input traffic features and weather information.
\subsection{Relation Mining in Multisource Data}
Relations in multisource data are primarily presented in the form of networks, and mining the structural and relational information contained in networks through representation vectors becomes the main approach to capture network information. In general, networks can be divided into homogeneous networks and heterogeneous networks according to the types of nodes. Homogeneous networks only consider one type of data, that is, the types of nodes must be identical; however, the majority of real networks have different types of nodes. To overcome the limited expressiveness of homogeneous networks, heterogeneous networks are proposed to represent the information of different types of nodes and the relationships between nodes. PTE \cite{tang2015pte} classifies texts, words, and labels and represents their pairwise relations to construct heterogeneous networks. \cite{gui2016large} and \cite{gui2017embedding} propose the HEBE embedding framework, which models the events with strong correlations as a whole to construct heterogeneous networks of events. A major drawback of the aforementioned heterogeneous networks is that accurate metapaths should be constructed when representing the relations between nodes, and specific metapaths may cause heterogeneous networks to be restricted to the framework of a particular network. In recent years, the emergence of knowledge graphs has provided broader ideas for the above problem. The modern concept of knowledge was first proposed by Google and then has been applied in various fields. Because of the power of knowledge graphs in processing graphical structures and information, increasingly more researchers have begun to understand and apply knowledge graphs in various fields, such as social networks \cite{noy2019industry}, search engines \cite{kasneci2008naga}, intelligent Q\&A systems and intelligent recommendations \cite{sha2019attentive}. Knowledge graphs are also applied in industries such as e-commerce \cite{wang2019multi}. They also play roles in transportation, such as in site selection \cite{shan2017follow} and traffic accidents \cite{xu2016building, muppalla2017knowledge}.

\section{Methods}
\subsection{Framework}
\begin{figure}[ht]
\centering
\includegraphics[width=1\linewidth]{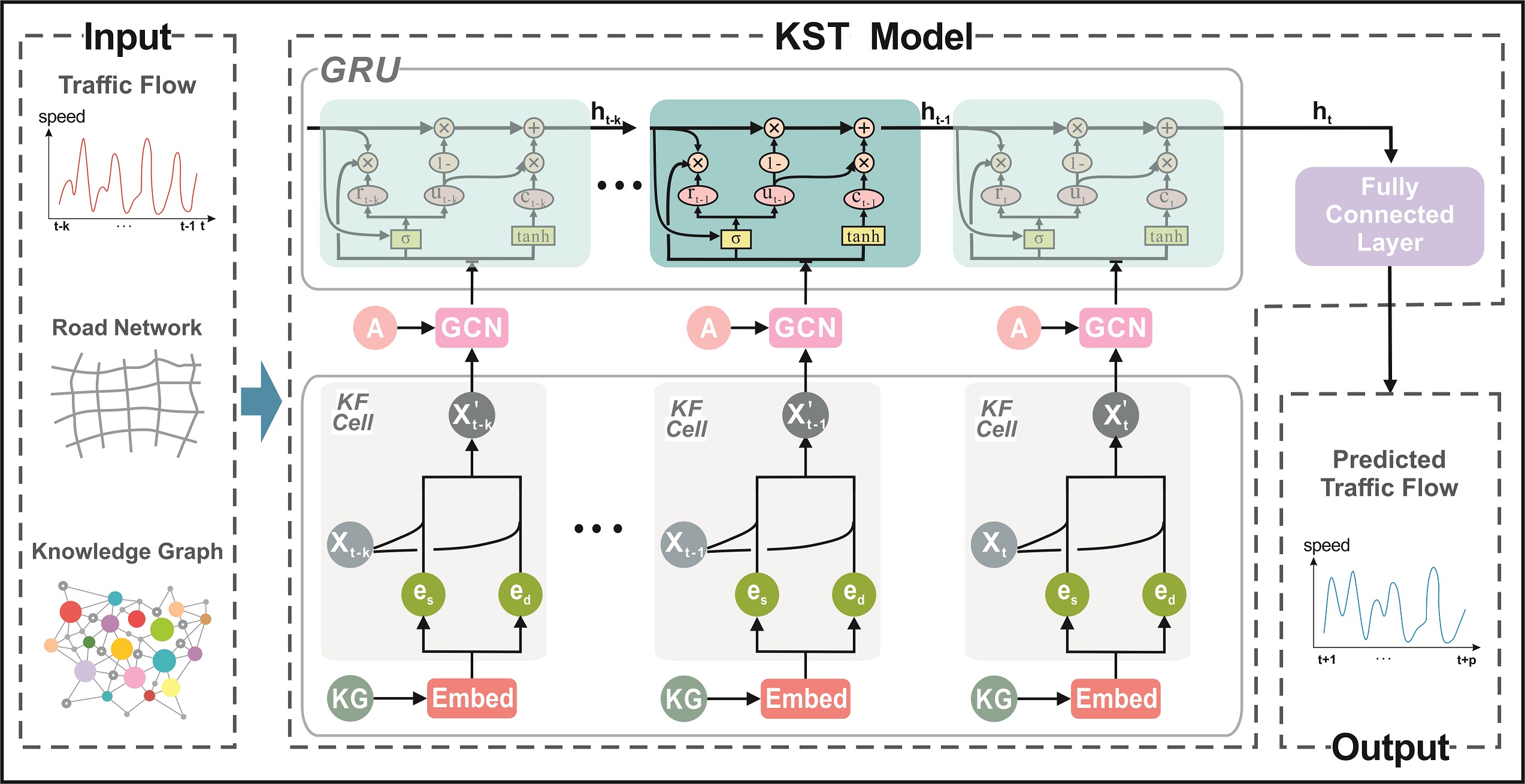} 
\caption{The framework of the KST-GCN.}
\label{fig1}
\end{figure}

In order to take into account the correlation between traffic information and external factors when predicting traffic flows, we first construct a specially designed knowledge graph, then use a knowledge graph representation learning model to derive embeddings encoded with knowledge of correlations. Later,  to incorporate the backbone spatial-temporal graph convolution network with knowledge, the embeddings and traffic features are fused by the KF-Cell before feeding into the spatial-temporal graph convolution network. By this means, the knowledge-driven spatial-temporal graph convolutional network (KST-GCN) is able to capture the spatial-temporal features of traffic in addition to the knowledge structures and semantic relations between traffic information and attributes. The workflow of the KST-GCN is shown in Figure \ref{fig1}.

To sum up, the traffic forecasting problem can be considered as learning the function $f$ to calculate the traffic characteristics of a future period given the traffic network topology $A$, the feature matrix $X$, and the traffic knowledge graph $CKG$:
\begin{equation}
y=f(A,X,CKG),  \label{equa:1}
 \end{equation}
where $A$ is the adjacency matrix of a road network, the value of the entry $a_{ij}$ in $A$ is set to 1 if the corresponding road section $i$ and road section $j$ are connected and 0 otherwise. $X$ represents the features of each node in the urban road network, and we choose the traffic speed as the node feature in this paper. For more details on city knowledge graph $CKG$, refer to Sections\ref{KR} and \ref{KG}.
\subsection{Knowledge Graph Representation Learning}
\subsubsection{Knowledge Graph}
\begin{figure}[ht]
  \centering
  \includegraphics[width=0.6\linewidth]{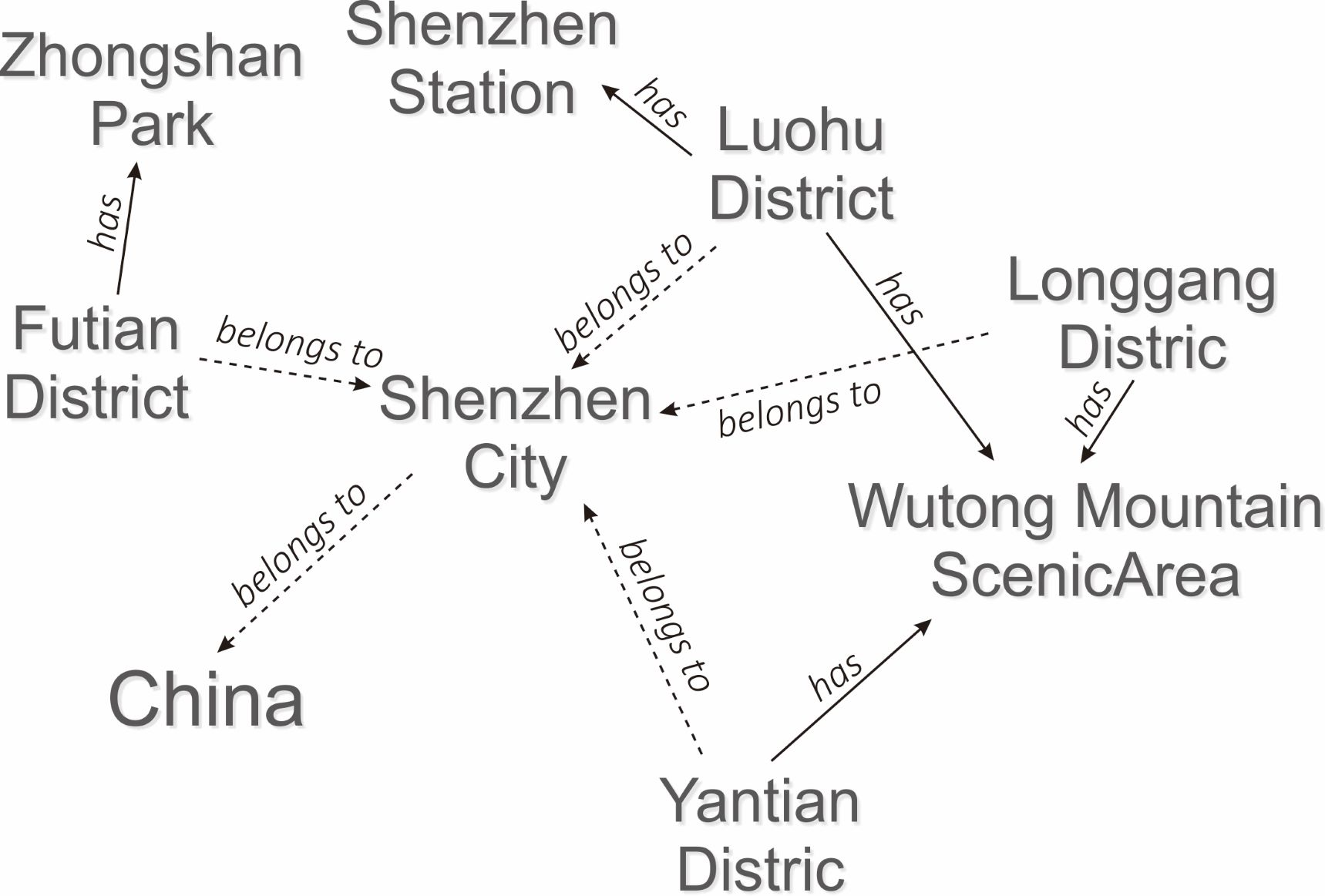} 
  \caption{An illustrative example of a knowledge graph.}
  \label{fig2}
\end{figure}

A knowledge graph, which enables the fusion of data from various sources while preserving the original information, can be defined as a knowledge network composed of multiple triples (head, relation, tail) with semantic information and network structures \cite{chen2020review}. The head and the tail in a triple are both entities, and the relation is a semantic relation between entities. A knowledge graph can represent heterogeneous nodes and multi-relational information. An illustrative example of a knowledge graph is shown in Figure \ref{fig2}. Hierarchical and semantic relations between the concepts can be discovered from the figure. 
\subsubsection{Knowledge Representation}\label{KR}
Most knowledge graphs (KGs) use entity-relation-based representations, where relation in the triple is used to model both attributes of entities and the relationship between entities. However, a relation is a semantic relationship between two entities, and an attribute is a property of an entity itself; the two concepts are fundamentally different. Besides, attributes and entities generally have one-to-many, many-to-one, or many-to-many relationships. The relationships between attribute factors and the traffic road network are many-to-one and one-to-many. Thus, in principle, knowledge graph representations that distinguish attribute and relational information are more suitable for capturing semantic information and correlations in this scenario. Therefore, we adopt the entity-attribute-relationship-based knowledge graph representation model KR-EAR \cite{Lin2016knowledge} to capture the knowledge structure and semantic information between road sections and external factors.
\begin{figure}[ht]
\centering
\includegraphics[width=0.9\linewidth]{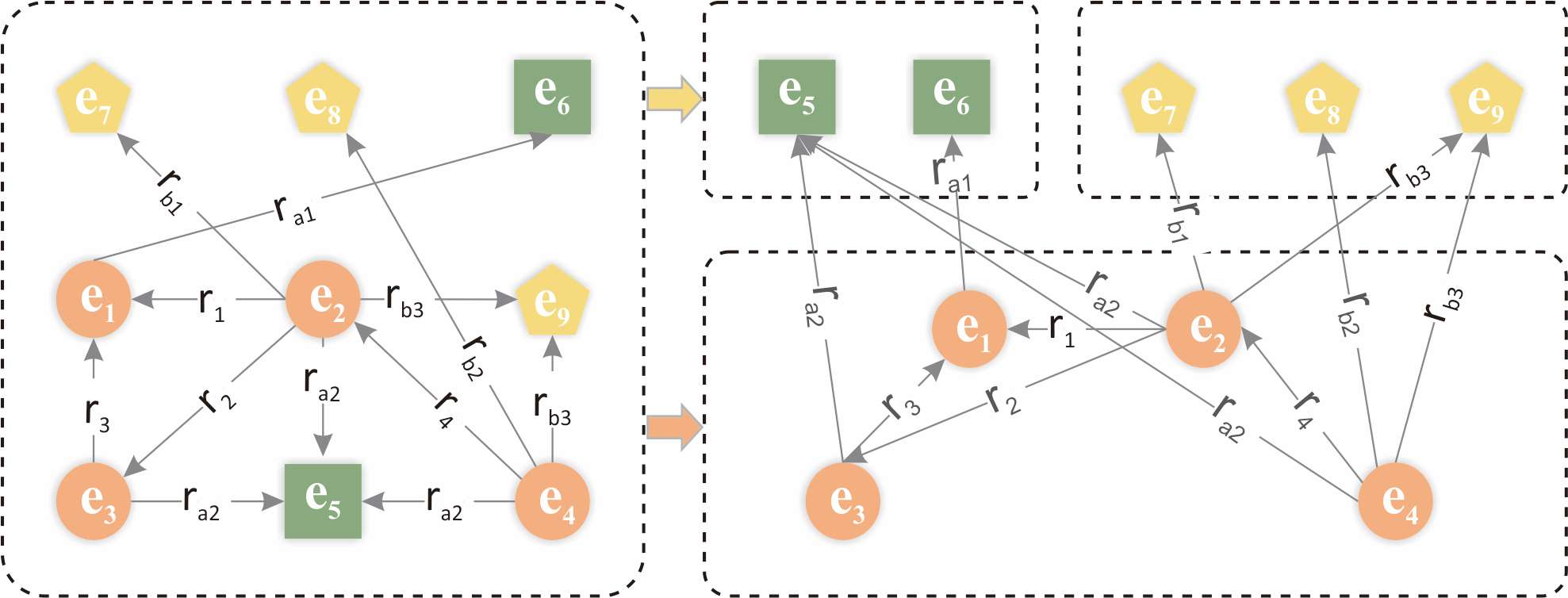} 
\caption{Entity-relationship-based representation method (left) and entity-attribute-relationship-based representation method (right). The circles indicate entities while the squares and pentagons indicate the values of two different types of attributes.}
\label{fig3}
\end{figure}

Specifically, as shown in Figure \ref{fig3}, the left dashed box is the knowledge graph constructed based on the entity-relationship-based representation, and the right dashed box is the entity-attribute-relationship-based knowledge graph representation adopted by KR-EAR. $r_i$ is the $ith$ class of relationships in the knowledge graph, where $r_a$ and $r_b$ are the attribute classes that connect the two classes of attribute values to entities. In the entity-relationship-based model, all triples are modeled for learning representations, while in the entity-attribute-relationship-based model, the triples are split into attribute triples and relation triples (as shown in the upper and lower parts on the right side of Figure \ref{fig3}). For example, $(e_1, r_{a1}, e_6)$ is an attribute triple and $(e_2, r_2, e_3)$ is a relation triple.

In this paper, roads, attributes, and the relationships between them are expressed in the form of triples of $CKG = \{R, R \_ att, att \_ att\}$.
\begin{equation}
R=\left\{\left(v_{i}, a d j, v_{j}\right), i, j \in\{1,2, \ldots, n\}\right\}.  \label{equa:2}
 \end{equation}
As shown in Equation \ref{equa:2}, $R$ is the relation triple that represent the adjacency relation $adj$ between road section $v_i$ and $v_j$, $n$ is the number of road sections. 
\begin{equation}
\centering
\begin{aligned}
R_{-} a t t &=\{(v_{i}, a t t_{l}, a t t_{l-} v_{i})\}, 
\\i, j \in\{1,2, &\ldots, n\}, l \in\{1,2, \ldots, L\}\}.
\label{equa:3}
\end{aligned}
\end{equation}
$R\_att$ is the attribute triple that represents the correspondence between roads and attributes, which is formed as Equation \ref{equa:3}, where $att_l$ is the $lth$ class of attributes, $a t t_{l-v_{i}}$ is the corresponding attribute value (e.g., the weather is sunny) of road section $i$, and $L$ is the number of attribute categories. 
 \begin{equation}
att _{-} at t=\left\{\left(a t t_{l_{1}}, a t t_{l_{2}}, p\right), l_{1}, l_{2} \in\{1,2, \ldots, L\}\right\}. \label{equa:4}
\end{equation}
The co-occurrence relations between attributes $att\_att$ is defined as Equation \ref{equa:4}, where $att_{l_{1}}$ and $att_{l_{2}}$ denote two different attributes, and $p$ is their co-occurrence probability. 

Based on the entity-attribute-relationship knowledge graph, KR-EAR seeks to learn embeddings of entities, relations and attributes, which denoted as $X_E$. The objective function is then defined by maximizing the joint probability of relation triples and attribute triples given the embedding vectors $X_E$, which is formalized as:
\begin{equation}
\begin{aligned}
P(R, R{\_att} \mid  X_{E})&=P(R \mid X_{E}) P(R{\_att} \mid X_{E}),
\end{aligned}
\label{equa:5}
\end{equation}
\begin{equation}
\begin{aligned}
P(R \mid X_{E})=\\
\prod_{\left(v, a d j, v_{j}\right) \in R}& P\left(\left(v_{i}, a d j, v_{j}\right) \mid X_{E}\right),
\end{aligned}
\label{equa:6}
\end{equation}
\begin{equation}
\begin{aligned}
P(R{\_att}\mid X_{E} )  \quad&=\\ \prod_{(v_{i}, a t t_{l}, a t t_{l-} v_{i})\in R_{-} a t t}& P\left((v_{i}, a t t_{l}, a t t_{l-} v_{i}) \mid X_{E}\right),
\end{aligned}
\label{equa:7}
\end{equation}

where $P((v_i, adj, v_j) | X_E)$ denotes the conditional probability of the relation triple $(v_i, adj, v_j)$, $P((v_i, att, att_v) | X_E)$ denotes the conditional probability of the attribute triple $(v_i, att, att\_v)$, and $V = \{v_1,v_2,\cdots, v_n\}$ is a set of road section entities. The conditional probability of a relation triple is generated by TransR \cite{Lin15learningentity}, which can be defined as:
\begin{equation}
P\left(\left(v_{i}, a d j, v_{j}\right) \mid X_{E}\right)=\frac{\exp \left(g\left(v_{i}, a d j, v_{j}\right)\right)}{\sum_{\hat{v_{i}} \in V} \exp \left(g\left(\hat{v_{i}}, a d j, v_{j}\right)\right)},
\label{equa:8}
\end{equation}
\begin{equation}
g\left(v_{i}, a d j, v_{j}\right)=-\left\|v_{i} M_{r}+a d j-v_{j} M_{r}\right\|_{L 1 / L 2}+b_{1}.\label{equa:10}
\end{equation}
$g\left(v_{i}, a d j, v_{j}\right)$ is the energy function, which suggests the correlations of relations and entity pairs. \ref{equa:10} is the objective functions of TransR, where $b_1$ is a bias term, and $M_r$ stands for the transfer matrix, $L 1$ and $L 2$ represent the $L_1$ and $L_2$ norm. 

Meanwhile, a classification model is applied to capture the correlation between entities and attributes for attribute triple encoding. The equation for calculating the conditional probability of an attribute triple in the objective function can be defined as:
\begin{equation}
\begin{aligned}
P((v_{i}, & a t t_{l}, a t t_{l-} v_{i})\mid X_{E})=\\ &\frac{\exp \left(h\left(v_{i}, a t t_{l}, a t t_{l-} v_{i}\right)\right)}{\sum_{\hat {a t t_{l-} v_{i}} \in a t t_{-} v} \exp \left(h\left(v_{i},  a t t_{l}, \hat {a t t_{l-} v_{i}}\right)\right)},\label{equa:11}
\end{aligned}
\end{equation}
\begin{equation}
\begin{aligned}
h(v_{i}, & a t t_{l}, a t t_{l-} v_{i})=\\ &-\left\|f\left(v_{i} W_{a t t}+b_{a t t}\right)-E_{a t t\_{v}}\right\|_{L_{1 / L 2}}+b_{2},\label{equa:12}
\end{aligned}
\end{equation}

where $f$ is a nonlinear function (e.g., $tanh$),  $att\_v$ is the value set of attributes, $E_{att\_v}$ is the embedding vector of attribute value $att\_v$,  $W_{a t t}$ is a linear transformation, and $b_2$, $b_{a t t}$ are bias terms. By this means, KR-EAR generates representations for relations and attributes separately while strengthening the correlations between attributes. 
The computational complexity of KR-EAR is $(S_1 + S_2) K^2$, where $K$ is the embedding dimension, $S_1$ is the number of relational triples and $S_2$ is the number of attribute triples. 

\subsection{KF-Cell}
\begin{figure}[ht]
\centering
\includegraphics[width=0.8\linewidth]{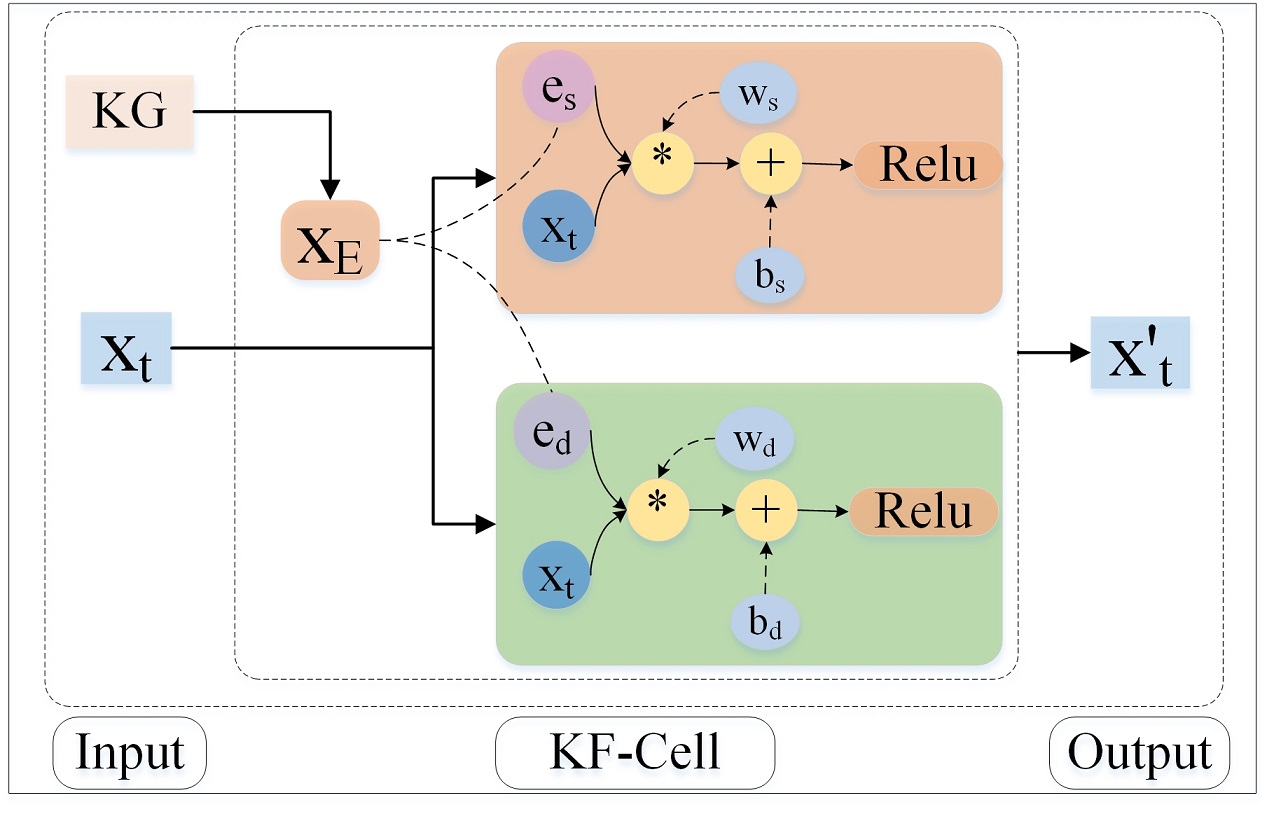} 
\caption{KF-Cell.}
\label{fig4}
\end{figure}

To perceive the knowledge of external factors as well as the correlations between the factors, and to model the spatial-temporal dependence of traffic flows based on the derived knowledge, we design the Knowledge Fusion Cell (KF-Cell), which introduces external knowledge into the backbone spatial-temporal graph convolutional networks. The KF-Cell is adaptable to the traffic forecasting models based on Graph Neural Networks and Recurrent Neural Networks without loss of generality. The details of the KF-Cell are shown in Figure \ref{fig4}.
The inputs to KF-Cell are knowledge embeddings $X_E$, and road section features $X_t$ observed at time $t$. The output is the updated road section features fused with external knowledge at time $t$, which we denoted as $X_t'$. Due to the diversity of external factors, we divided the external factors into two categories: the static factors and the dynamic factors. $e_s$ and $e_d$ in Figure \ref{fig4} and Equation \ref{KF} denote the road section embeddings with regards to the static and dynamic external factors, respectively, after applying KR-EAR. $w_d$, $w_s$ are linear transformations, $b_d$, $b_s$ are bias constants, and $[\cdot ]$ is the concatenation.
\begin{equation}
\begin{aligned}
X_s = Relu &(e_s x_t w_s + b_s)\\
X_d = Relu &(e_d x_t w_d + b_d)\\
X_t' = &[X_s, X_d].
\label{KF}
\end{aligned}
\end{equation}
To model the spatial dependence of a traffic flow based on knowledge representation, we use the updated road section features $X_t'$ and the adjacency matrix $A$ as the input of the GCN. We prefer GCN because it is more suitable for non-Euclidean structured data such as traffic networks than CNN. Furthermore, it is based on the idea that each road section in the network is influenced by its own and its upstream and downstream road sections. The GCN uses graph spectral  theory to capture the topological relations  and features of the traffic network and obtain the representation vector of each road section:
\begin{equation}
gc(y_l',A)=\sigma (\tilde{D}^{-\frac{1}{2}}\tilde{A}\tilde{D}^{-\frac{1}{2}}y_l'{{W}_{l}}),
\end{equation}
where $\sigma(*)$ is an activation function, $A$ is the adjacency matrix, $\tilde{A}$ is the adjacency matrix with self-connections, $\tilde{D}$ is the degree matrix of $\tilde{A}$, ${W}_{l}$ is the weight matrix of the $l-th$ convolution layer, and $y_l'$ is the output of the nonlinear combination of node features of the $l-th$ layer. In the first layer, $y_l'$ has an initial value of updated feature matrix $X_t'$.

Then, the recurrent neural networks are leveraged to capture the temporal dependency. In particular, we use Gated Recurrent Units (GRU)  as an example to show the whole workflow of the knowledge-driven traffic forecasting. The GRU model consists of a reset gate and an update, which is formulated as  Equation \ref{equa:14}. 
\begin{equation}
\begin{aligned}
u_{t}=\sigma(W_{u}g c &([X_t', h_{t-1}], A)+b_{u})\\
r_{t}=\sigma(W_{r}g c &([X_t', h_{t-1}], A)+b_{r})\\
c_{t}=\tanh  (W_{c}g c(&[X_t',(r_{t} \odot h_{t-1})], A)+b_{c})\\
h_{t}=u_{t}{\odot}&h_{t-1}+(1-u_{t}){\odot} c_{t},
\label{equa:14}
\end{aligned}
\end{equation}
where $[\cdot ]$ represents concatenation, $\odot$ is the Hadmard product. $r_t$ is the reset gate, which combines the information in the memory and the information at the current time step. $u_t$ is the update gate, which can select or forget memories. $\sigma$ is the signal used to represent the gated signal, and $c_t$ is the moment state at the current time step. In the memory updating phase, when $u_t$ acts as a forget gate, $u_t\odot h_{t-1}$ forgets unimportant previous information to update the memory; and when $u_t$ acts as a memory gate, $(1-u_t)\odot c_t$ remembers important information of the current. $W_{u},W_{r},W_{c}$ and $b_{u},b_{r},b_{c}$ are weights and bias, respectively. State $h_t$ represents the output at time $t$.

Finally, $h_t$ is fed into a fully connected layer to generate the predicted future speed $\hat{Y}$. The objective of the training process of KST-GCN is to minimize the error between the predicted traffic speed $\hat{Y}$ and real traffic speed $Y$; thus the loss function is formulated as:
\begin{equation}
loss=||Y-\hat{Y}||+\lambda L_{reg}.
\label{equa:15}
\end{equation}

\section{Experiments}
 \subsection{Data Description}
Although there are several open traffic flow datasets, it isn't easy to collect road network data, traffic data and additional knowledge information including POI data and weather data of the same area for the same period altogether. Limited by the data acquisition and the difficulty of constructing the knowledge graph, the experiments in this paper are all based on one dataset from Luohu District, Shenzhen. However, due to the generality of our experimental setup, experiments can be easily validated in other cities as long as the dataset is given.

The Shenzhen dataset contains taxi track data, the road network data, weather data, and POI data of each street from January 1 to January 31, 2015. The study area includes 156 road sections and nine types of POIs: food services, enterprises, shopping services, transportation services, education services, living services, medical services, accommodation services, and others. In addition, the weather data, including the temperature, weather conditions, wind speed, humidity, barometric pressure, and visibility, of the study area at 15 min intervals in January were crawled as auxiliary data. We classify weather conditions into five categories: sunny, cloudy, foggy, light rain, and heavy rain. 
\subsubsection{Knowledge Graph}\label{KG}
We count the number of POIs on each section and then use the road sections, categories, and numbers of POIs to construct attribute triples. For example, (road section 1, restaurant, 15) and (road section 1, school, 6) indicate that there are six schools and 15 restaurants around road section 1. Time, weather conditions, and their correlations, such as (road section ID, weather condition, moment) and (moment t, weather, light rain), are also used to construct the knowledge graph. Besides, the input data need to be preprocessed before being input into the knowledge graph embedding model. The input data include the road adjacency triples (head entity, relation, tail entity), attribute triples (entity, attribute, attribute value), and attribute co-occurrence triples (attribute 1, attribute 2, co-occurrence probability). The attribute co-occurrence probability describes the probability of two attributes existing on the same road section. The data structures of these triples are shown in Table \ref{tbl:t1}, Table \ref{tbl:t2}, and Table \ref{tbl:t3}.
\begin{table}[htbp]
\centering
\caption{Road Adjacency Triples}
\begin{tabular}{ccc} 
\hline
Head Entity & Relation & Tail Entity  \\ 
\hline
90217       & adj      & 95968        \\
...         & ...      & ...          \\
90225       & adj2     & 95504        \\
\hline
\end{tabular}
\label{tbl:t1}
\end{table}
\begin{table}[htbp]
\centering
\caption{Attribute  Triples}
\begin{tabular}{ccc} 
\hline
Head Entity & Attribute & Attribute Value  \\ 
\hline
90217       & transportation service      & 4        \\
...         & ...      & ...          \\
90217       & food services     & 31        \\
\hline
\end{tabular}
\label{tbl:t2}
\end{table}
\begin{table}[htbp]
\centering
\caption{Attribute Co-occurrence Triples}
\begin{tabular}{ccc} 
\hline
Attribute & Attribute & Co-occurrence Possibility  \\ 
\hline
transportation service       & food service      &   0.016      \\
...         & ...      & ...          \\
shopping service      & food service     & 0.255        \\
\hline
\end{tabular}
\label{tbl:t3}
\end{table}

\begin{figure}[htbp]
\centering
\includegraphics[width=0.85\linewidth]{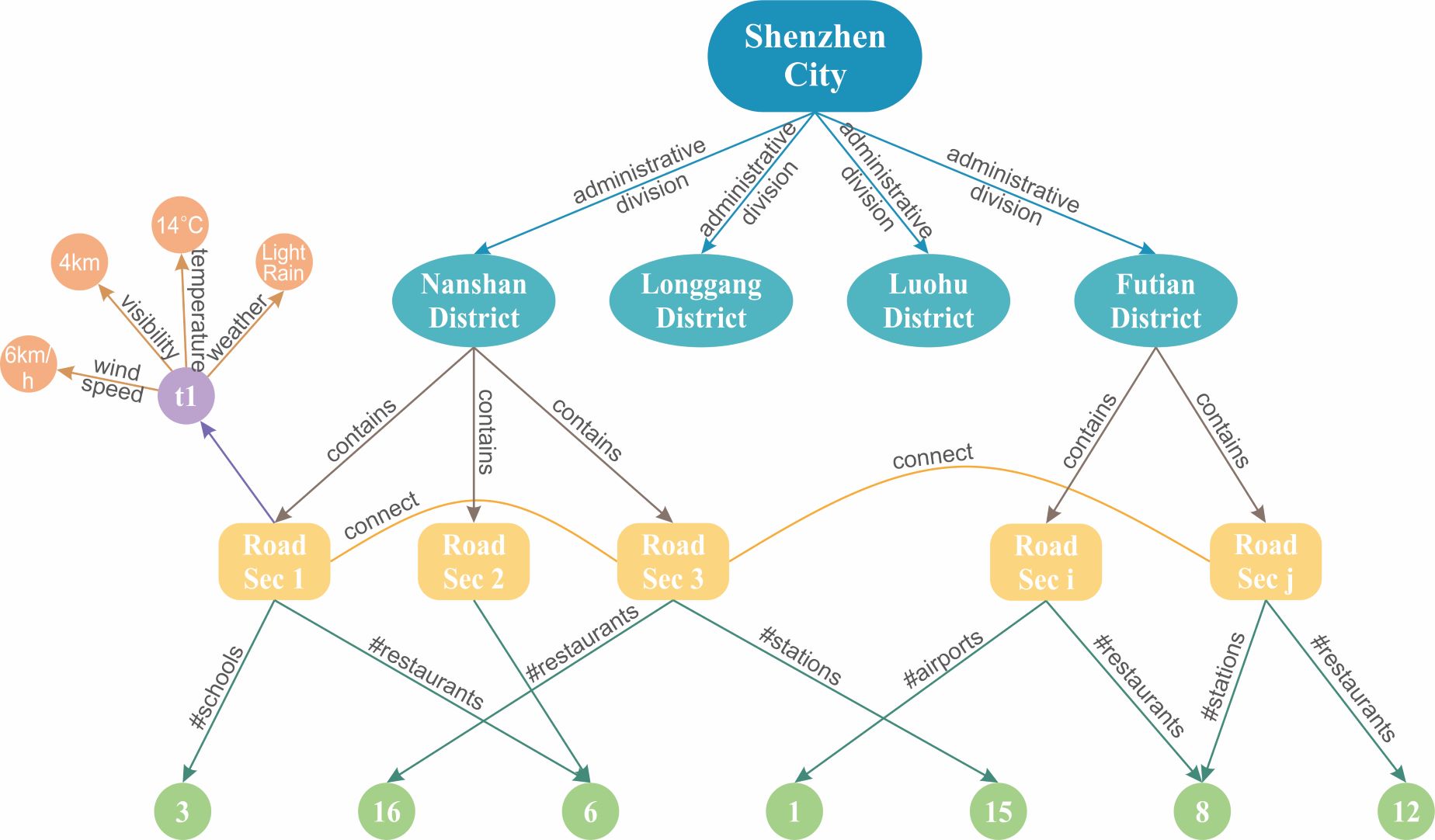} 
\caption{The city knowledge graph of Shenzhen.}
\label{fig5}
\end{figure}

Part of the final Shenzhen city knowledge graph is shown in Figure \ref{fig5}.

 \begin{figure}[htbp]
  \centering
  \subfloat[]{
  \includegraphics[width=0.4\textwidth]{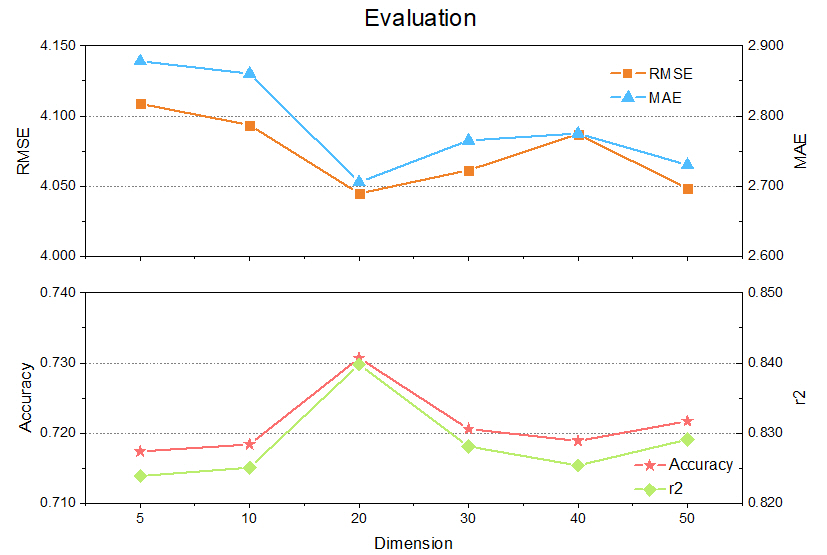}
  }\\
  \subfloat[]{
  \includegraphics[width=.4\textwidth]{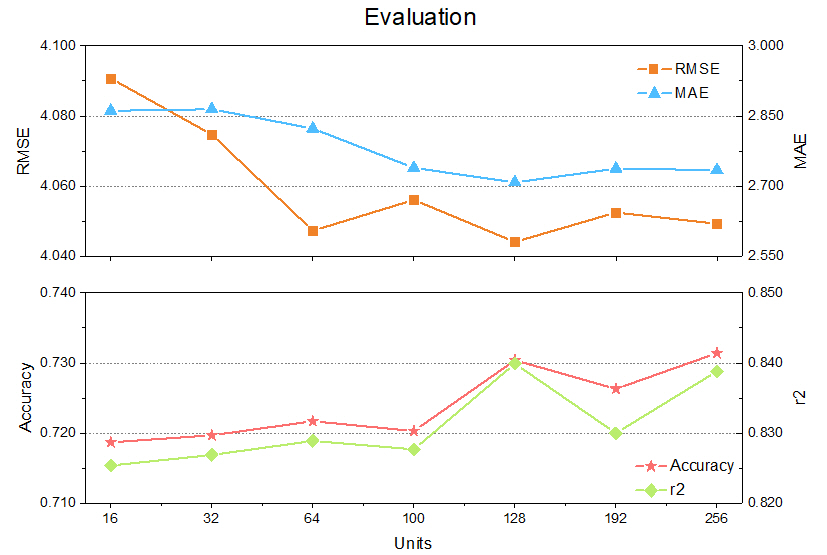}
  }
  \caption{Performance of KF-T-GCN under different hyperparameter settings} \label{fig6A}
  \end{figure}
 \begin{figure}[htbp]
  \centering
  \subfloat[]{
  \includegraphics[width=0.4\textwidth]{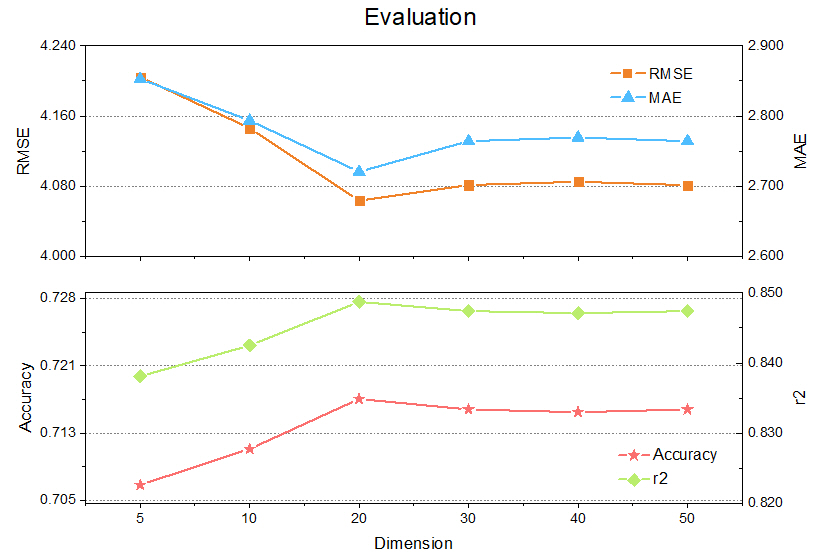}
  }\\
  \subfloat[]{
  \includegraphics[width=0.4\textwidth]{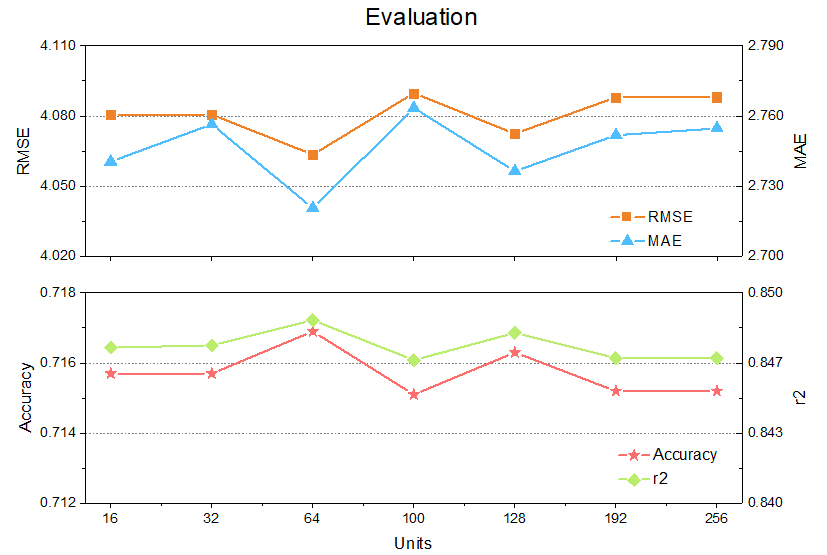}
  }
  \caption{Performance of KF-DCRNN under different hyperparameter settings} \label{fig6B}
  \end{figure}  
  
\subsection{Experimental Settings}
The objective of our method is to boost the traffic forecasting performance of the backbone models with the knowledge graph. Since the KF-Cell we designed to fuse knowledge is adaptable to the traffic forecasting models based on Graph Neural Networks and Recurrent Neural Networks without loss of generality, we evaluate the proposed method on two representative spatial-temporal graph convolutional networks, DCRNN \cite{yu2017spatiotemporal}, and T-GCN \cite{zhao2019t}, and name their corresponding KST-GCN variants as KF-DCRNN and KF-T-GCN, respectively. The task is to predict the future traffic speeds given historic traffic speeds, the underlying road networks, and the knowledge graph.

Based on experience, the learning rate is set to 0.001, and the proportion of the dataset used for training is 80\%. The number of hidden units and the knowledge embedding dimension are two important hyperparameters for our model, as they have the most significant impacts on the prediction results. Therefore, we conduct experiments to choose the appropriate values for these two hyperparameters: \textbf{(1) Embedding dimension}: we choose the embedding dimension from [5, 10, 20, 30, 40, 50] to analyze the model performances. Figure \ref{fig6A}(a) and Figure \ref{fig6B}(a) show the RMSE, MAE, accuracy, and $R^2$ of the models with different embedding dimension settings. It can be seen that all the KST-GCNs have the best performance when the embedding dimension is set to 20; \textbf{{(2) The number of hidden units}}: we fix the embedding dimension to 20 and choose the number of hidden units from [16, 32, 64, 128, 192, 256]. Figure \ref{fig6A}(b) and \ref{fig6B}(b) shows the RMSE, MAE, accuracy, and $R^2$ of the models with different numbers of hidden units. When the number of hidden units is set to 128, the KF-T-GCN has the best performance. As for the KF-DCRNN, the optimal choice is 64, where the best performance occurs.
\subsection{Results and Analysis}

\begin{table*}[htbp]
\centering
\caption{Experimental Comparison with Baselines and Backbones}

\begin{tabular}{ccccccccc} 
\hline
Evaluation matrix & SVR    & ARIMA  & GCN    & GRU    & DCRNN  & KF-DCRNN & T-GCN   & KF-T-GCN  \\ 
\hline
RMSE              & 7.2203 & 6.7708 & 5.6419 & 5.0649 & 4.1243 & 4.0635   & 4.0696 & 4.0443    \\
MAE               & 4.7762 & 4.6656 & 4.2265 & 2.5988 & 2.7514 & 2.7206   & 2.7460 & 2.7090    \\
Accuracy          & 0.706  & 0.3852 & 0.6119 & 0.7243 & 0.7127 & 0.7169   & 0.7165 & 0.7306    \\
r2                & 0.8367 & *      & 0.6678 & 0.8322 & 0.8441 & 0.8487   & 0.8388 & 0.8400    \\
var               & 0.8375 & 0.0111 & 0.6679 & 0.8322 & 0.8441 & 0.8491  & 0.8388 & 0.8400    \\
\hline
\end{tabular}
\label{tbl1}
\end{table*}
\subsubsection{Prediction Accuracy}
We compare the performance of the KST-GCNs with those of the baseline methods and backbones with a prediction horizon of 15 minutes. Table \ref{tbl1} shows that the KST-GCNs and their backbones all have better performances than the traditional methods like SVR and ARIMA and deep learning methods like GCN and GRU, the former cannot fit complex cases, while the latter only considers spatial or temporal dependencies alone.  The KST-GCNs all outperform their backbones, with the KF-DCRNN and KF-T-GCN reducing the RMSE by 1.47\% and 0.63\% over the DCRNN and T-GCN, respectively, verifying that the semantic information between roads and attributes boosts the prediction. 
\subsubsection{Long-term Prediction}\label{Long-term}
\begin{table}[htbp]
  \centering
  \caption{Performance under Different Prediction Horizons}
\resizebox{\linewidth}{!}{\begin{tabular}{cccccc} 
\hline
Time                   & Metric   & DCRNN  & KF-DCRNN & T-GCN   & KF-T-GCN  \\ 
\hline
\multirow{5}{*}{15min} & RMSE     & 4.1243 & 4.0635   & 4.0696 & 4.0443    \\
                       & MAE      & 2.7514 & 2.7206   & 2.7460  & 2.7090    \\
                       & Accuracy & 0.7127 & 0.7169   & 0.7165 & 0.7206    \\
                       & r2       & 0.8441 & 0.8487   & 0.8388 & 0.8400    \\
                       & var      & 0.8441 & 0.8491   & 0.8388 & 0.8400    \\ 
\hline
\multirow{5}{*}{30min} & RMSE     & 4.1669 & 4.0846   & 4.0770  & 4.0687    \\
                       & MAE      & 2.7902 & 2.7485   & 2.7470  & 2.7228    \\
                       & Accuracy & 0.7097 & 0.7154   & 0.7159 & 0.7201   \\
                       & r2       & 0.8409 & 0.8472   & 0.8377 & 0.8372    \\
                       & var      & 0.8409 & 0.8476   & 0.8377 & 0.8374    \\ 
\hline
\multirow{5}{*}{45min} & RMSE     & 4.1982 & 4.0979   & 4.1035 & 4.0775    \\
                       & MAE      & 2.8217 & 2.7628   & 2.7788 & 2.7698    \\
                       & Accuracy & 0.7075 & 0.7145   & 0.7141 & 0.7195    \\
                       & r2       & 0.8385 & 0.8462   & 0.8357 & 0.8365    \\
                       & var      & 0.8385 & 0.8466   & 0.8357 & 0.8365    \\ 
\hline
\multirow{5}{*}{60min} & RMSE     & 4.2288 & 4.1115   & 4.2660  & 4.0798    \\
                       & MAE      & 2.8412 & 2.7802   & 2.7911 & 2.7768    \\
                       & Accuracy & 0.7053 & 0.7135   & 0.7125 & 0.7194    \\
                       & r2       & 0.8361 & 0.8452   & 0.8339 & 0.8363    \\
                       & var      & 0.8361 & 0.8457   & 0.8340  & 0.8364    \\
\hline
\end{tabular}}\label{tbl2}
  \end{table}
We further compare the performance between the KST-GCNs and their backbones under various prediction horizons; the results are shown in Table \ref{tbl2}. We can observe that the performance of all models deteriorates with the increase of the prediction horizon. However, the KST-GCNs still outperform their backbones under each prediction horizon. As the prediction time increases, the performance of DCRNN gradually becomes worse, and KF-DCRNN increases its improvement with the RMSE reduction from 1.50\% to 2.85\% and Accuracy increment from 0.59\% to 1.15\%. The same pattern appears in the comparison between KF-T-GCN and T-GCN. The improvement of KF-T-GCN starts with the RMSE reduction of 0.62\% and ends with 4.36\% and starts with an Accuracy increment of 0.57\% and ends with 0.97\%. Therefore, it is safe to conclude that the KST-GCNs can maintain superiority for short-term and long-term predictions compared to their backbones.

\subsubsection{Knowledge Representation}
\begin{figure}
\centering
\subfloat[]{
\includegraphics[width=0.32\textwidth]{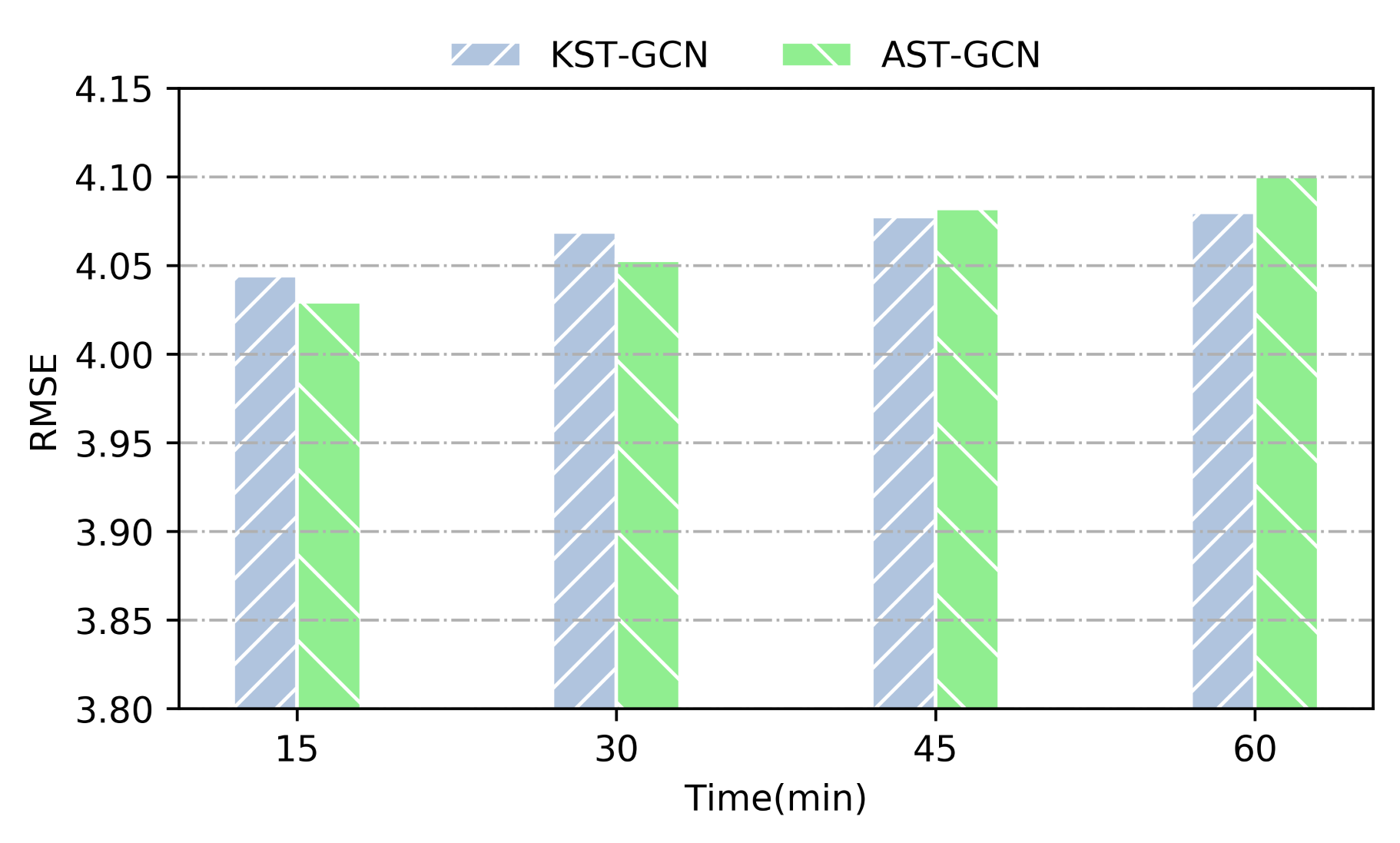}
}\\
\subfloat[]{
\includegraphics[width=.32\textwidth]{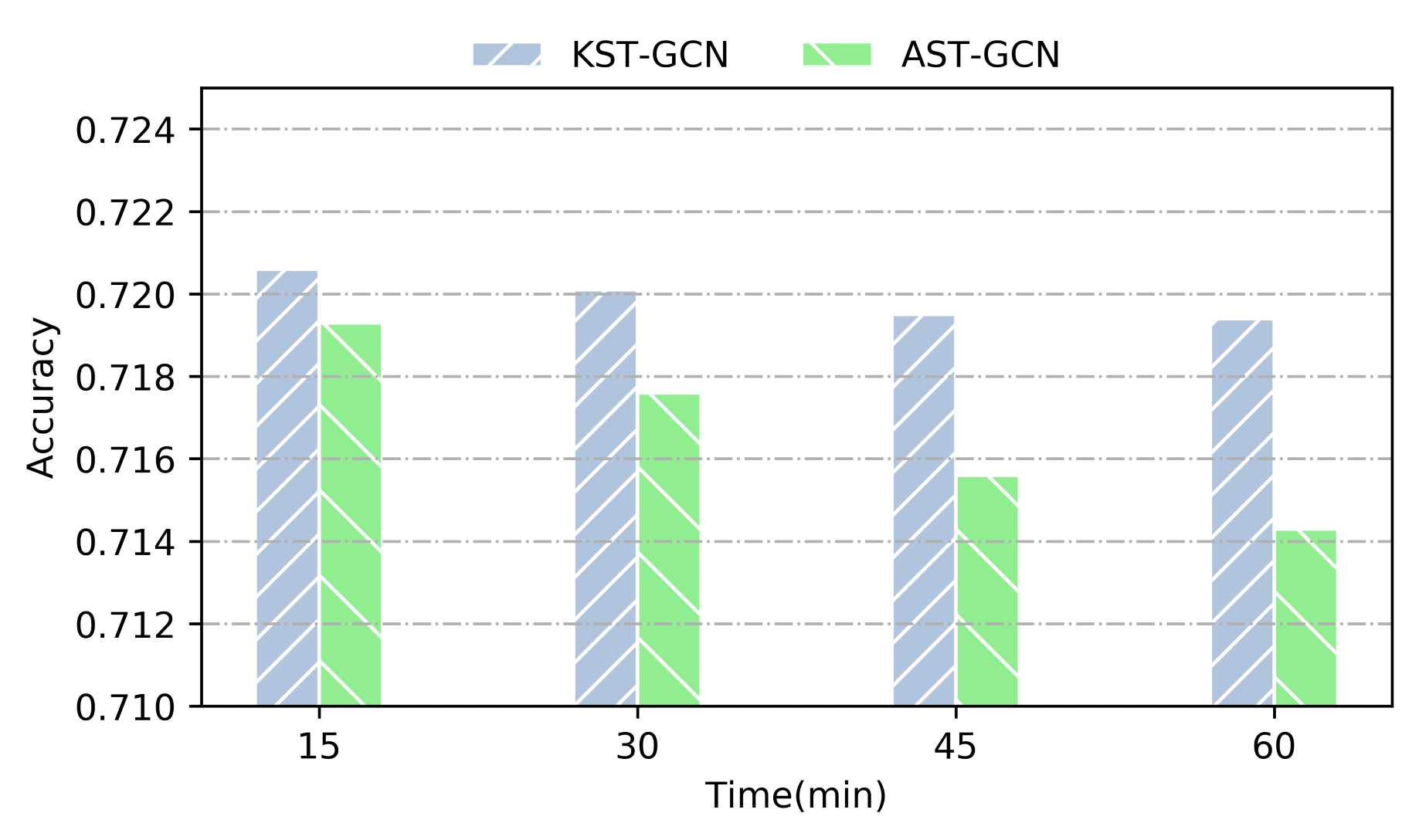}
}
\caption{Performance of KF-T-GCN and AST-GCN} \label{fig8}
\end{figure}

In this section, we verify the validity of the knowledge representation based on the knowledge graph. We compare the KF-T-GCN with the AST-GCN\cite{hf2020ast}, a model that enhances the feature matrix $X$ by directly concatenating external factor information without any knowledge translation. The following experiments analyzed the prediction errors and accuracies of the KF-T-GCN and AST-GCN under different prediction horizons. The results are shown in Figure \ref{fig8}.
Figure \ref{fig8} shows that the accuracy of the KF-T-GCN is higher than that of the AST-GCN at all prediction horizons, and the gap between the KF-T-GCN and AST-GCN increases as the prediction horizon increases. At a prediction horizon of 60 min, the KF-T-GCN outperforms the AST-GCN by 54.1\%. The RMSE of the KF-T-GCN, which represents the prediction error, is slightly higher at prediction horizons of 15 min and 30 min, but it is lower for long-term prediction and has a milder fluctuation. The above experimental results confirm the validity and stability incorporating the semantic knowledge in spatial-temporal graph convolutional network.
\subsubsection{Ablation Study}
We conduct ablation experiments to analyze the effects of incorporating different knowledge to spatial-temporal graph convolutional networks on the traffic prediction task. In the experiments, traffic features fused with POI knowledge, with weather knowledge, with the full knowledge, and without any additional attribute information are fed into models. The results are shown in Table \ref{tbl3} and \ref{tbl4}.

\begin{table}[htbp]
\centering
\caption{Ablation Study of KF-T-GCN}
\begin{tabular}{ccccc} 
\hline
\multirow{2}{*}{Metric} & \multirow{2}{*}{T-GCN} & \multicolumn{3}{c}{KF-T-GCN}  \\ 
\cline{3-5}
                        &                       & Weather & POI    & KG         \\ 
\hline
RMSE                    & 4.0696                & 4.0501  & 4.0489 & 4.0443     \\
MAE                     & 2.7460                & 2.7357  & 2.7428 & 2.7090      \\
Accuracy                & 0.7165                & 0.7215  & 0.7208 & 0.7206     \\
r2                      & 0.8388                & 0.8388  & 0.8381 & 0.8400       \\
var                     & 0.8388                & 0.8389  & 0.8381 & 0.8400       \\
\hline
\end{tabular}
\label{tbl3}
\end{table}
\begin{table}[htbp]
\centering
\caption{Ablation Study of KF-DCRNN}
\begin{tabular}{ccccc} 
\hline
\multirow{2}{*}{Metric} & \multirow{2}{*}{DCRNN} & \multicolumn{3}{c}{KF-DCRNN}  \\ 
\cline{3-5}
                        &                        & Weather & POI    & KG         \\ 
\hline
RMSE                    & 4.1243                & 4.0830  & 4.0805 & 4.0635     \\
MAE                     & 2.7514                 & 2.7423  & 2.7606 & 2.7206     \\
Accuracy                & 0.7127                 & 0.7156  & 0.7157 & 0.7169     \\
r2                      & 0.8441                & 0.8473  & 0.8475 & 0.8487     \\
var                     & 0.8441                 & 0.8477  & 0.8478 & 0.8491     \\
\hline
\end{tabular}
\label{tbl4}
\end{table}
It can be seen that the RMSE of the KF-T-GCN (weather) is 0.28\% lower than that of the T-GCN, the RMSE of the KF-T-GCN (POI) is 0.32\% lower than that of the T-GCN. Regarding fusing roads with both POI and weather knowledge, the KF-T-GCN (KG) has a 0.62\% lower prediction error than that of the T-GCN. 

From Table \ref{tbl4}, it can be seen that the RMSE of the KF-DCRNN (weather) is 1.00\% lower than that of the DCRNN, the RMSE of the KF-DCRNN (POI) is 1.06\% lower than that of the DCRNN. Regarding fusing roads with both POI and weather knowledge, the KF-DCRNN (KG) has a 1.47\% lower prediction error than that of the DCRNN. 

To sum up, the model that incorporates POI knowledge outperforms the model that incorporates weather knowledge structure, indicating that the semantic correlation between roads and POI is more pronounced in traffic prediction tasks. Overall, the fusion of roads and knowledge can assist the prediction model to some extent and improve the prediction accuracy.

\subsubsection{Robustness Analysis}
\begin{figure}[htbp]
\caption{Perturbation Analysis of KF-T-GCN}
\label{fig9a}
\centering
\subfloat[Gaussian perturbation]{\includegraphics[width=0.252\textwidth]{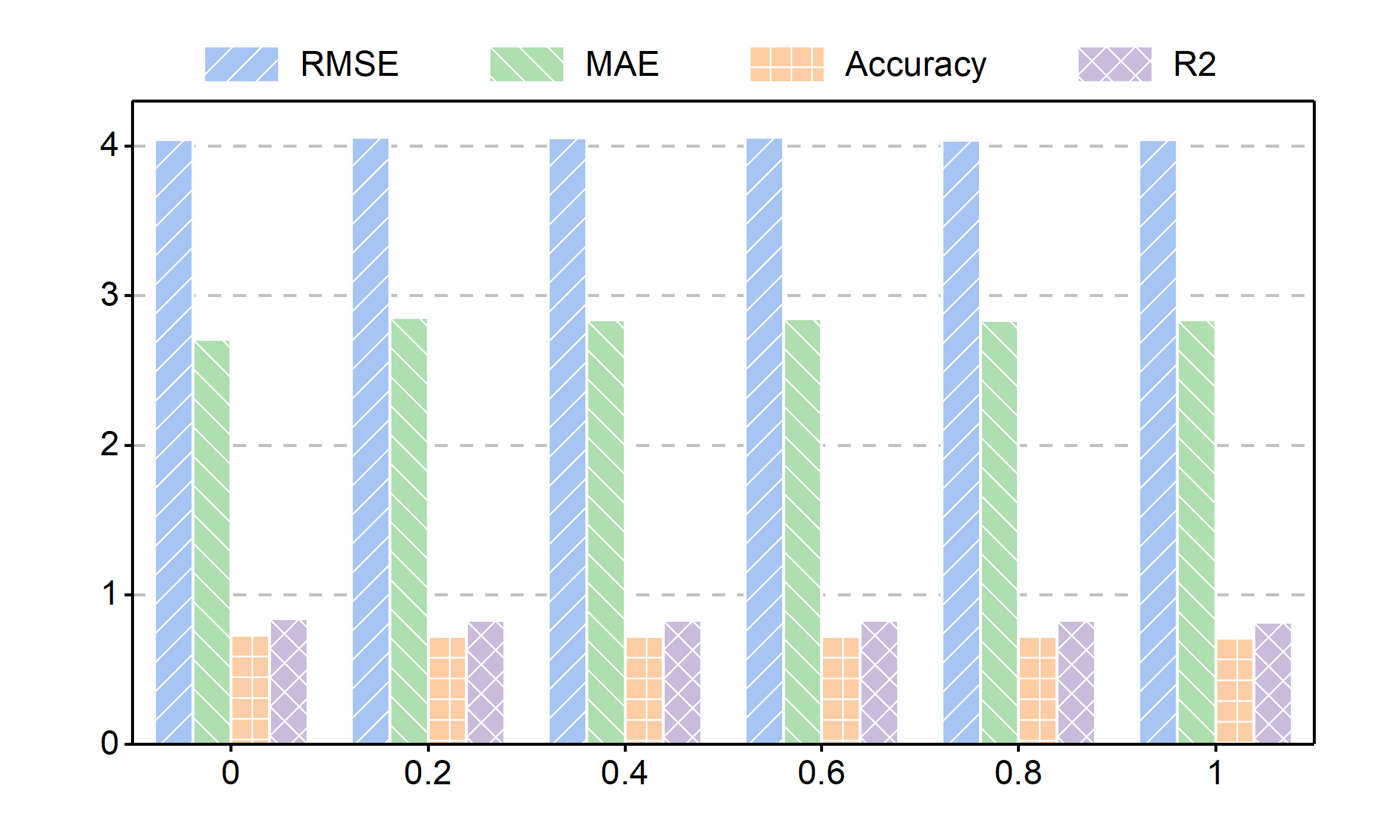}}
\subfloat[Poisson perturbation]{\includegraphics[width=0.252\textwidth]{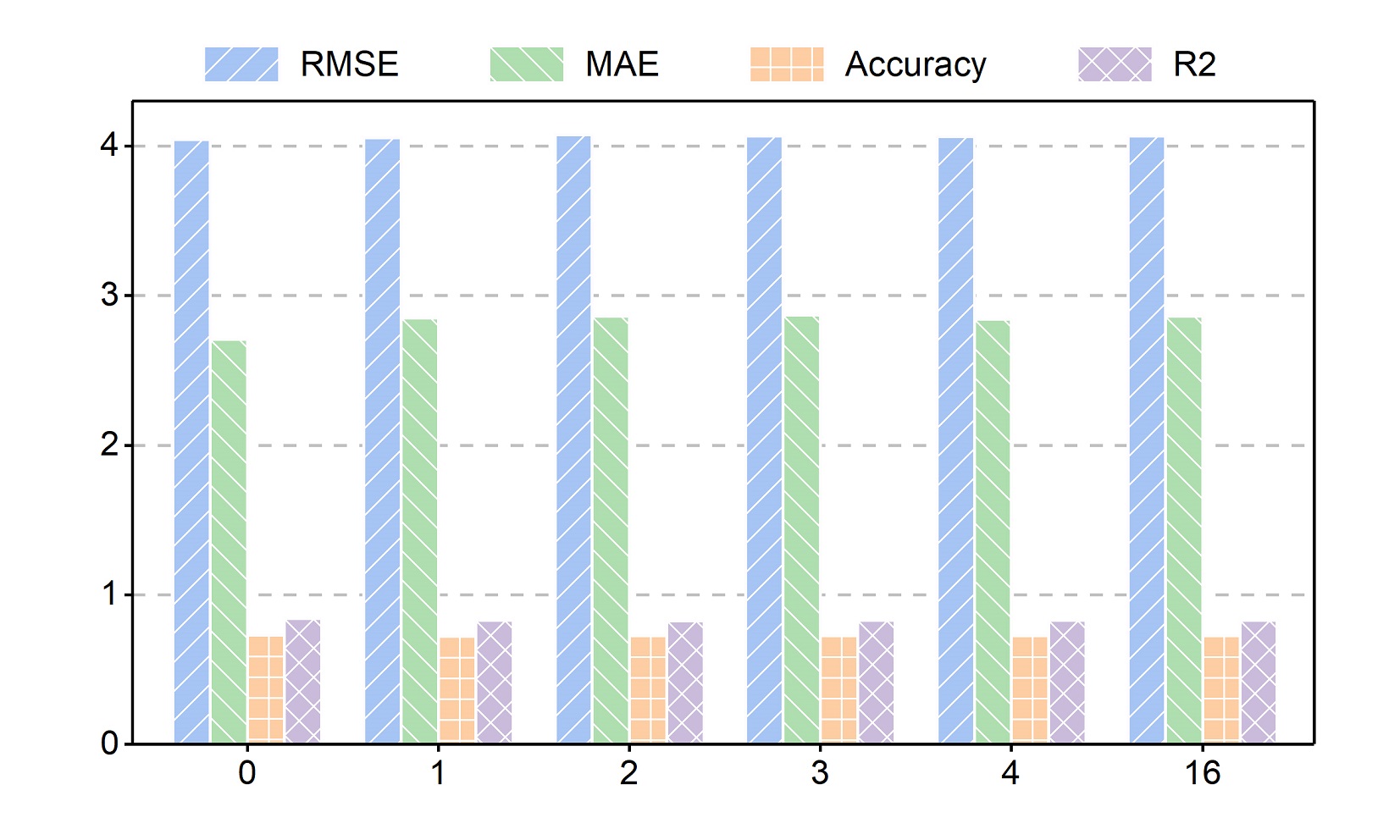}}
\end{figure}

\begin{figure}[htbp]
\caption{Perturbation Analysis of KF-DCRNN}
\label{fig9b}
\centering
\subfloat[Gaussian perturbation]{\includegraphics[width=0.252\textwidth]{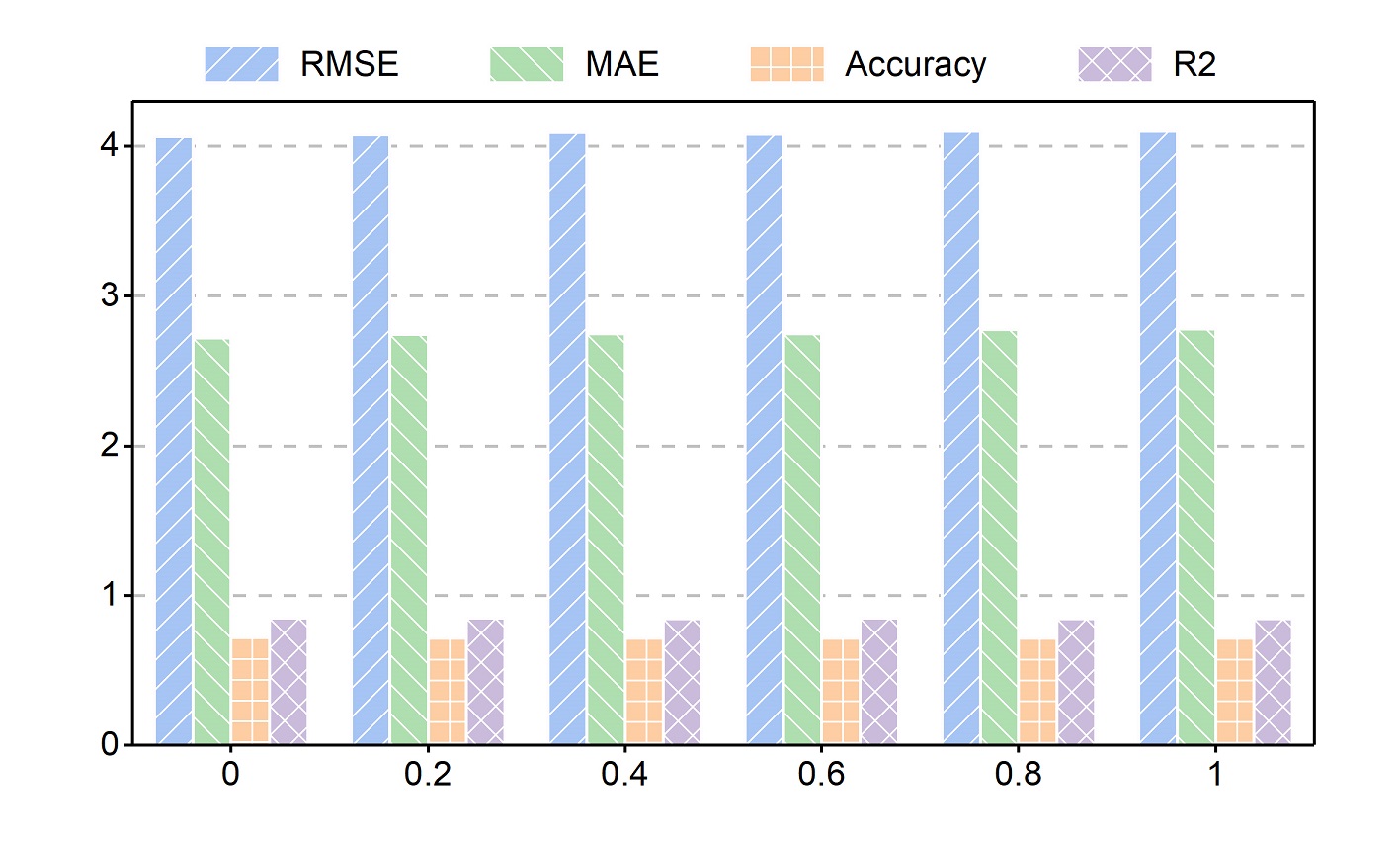}}
\subfloat[Poisson perturbation]{\includegraphics[width=0.252\textwidth]{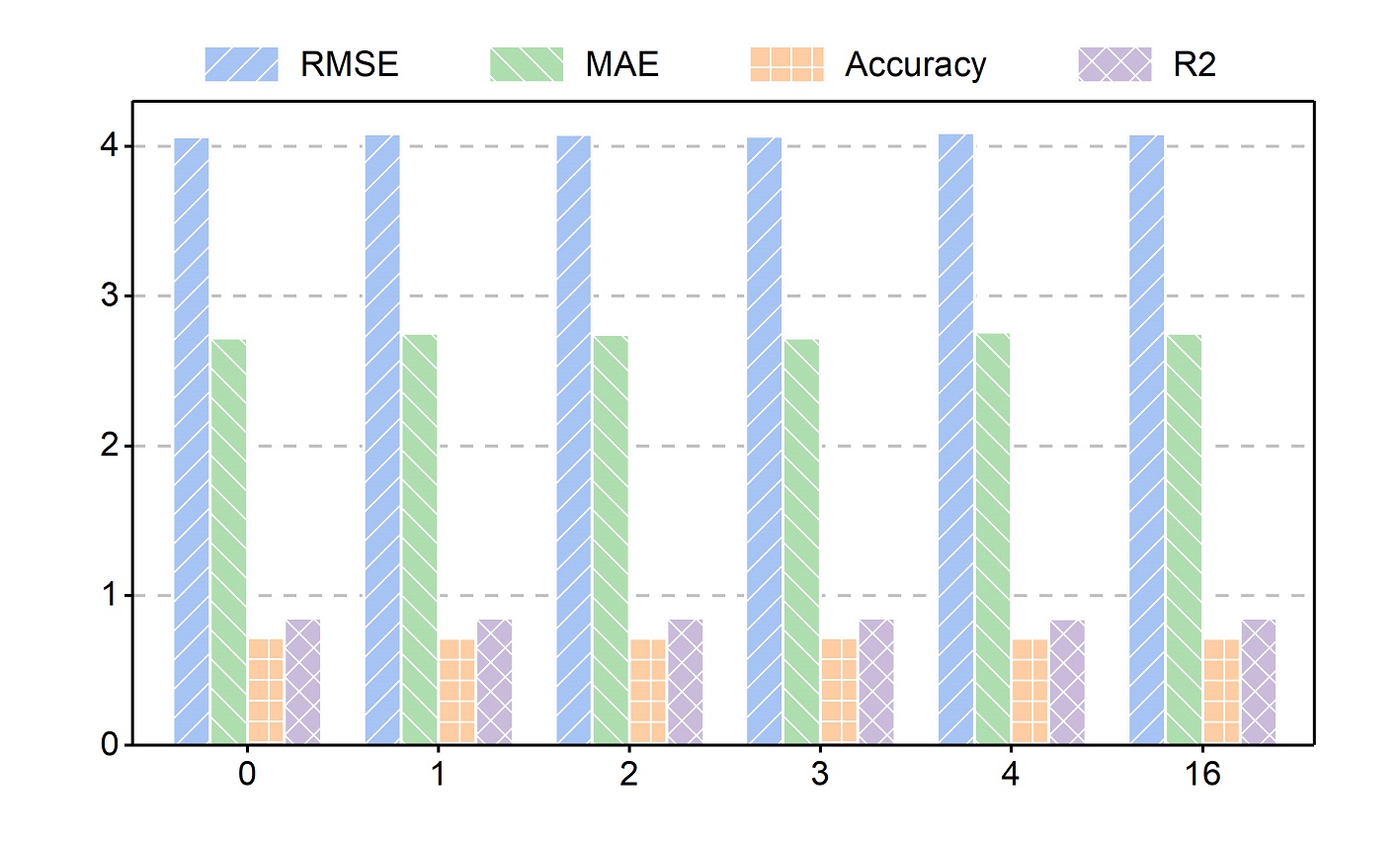}}
\end{figure}

Real-world urban data are rich in information but contain noise. To understand the effect of noise on the KST-GCN, Gaussian noise and Poisson noise are added to the original urban traffic data. The added noise obeys a Gaussian distribution $N \in (0, \sigma^2) (\sigma \in 0.2, 0.4, 0.6, 0.8, 1, 2)$ and a Poisson distribution $P(\lambda) (\lambda \in 1, 2, 4, 8, 16)$. The experimental results after adding noise are shown in Figure \ref{fig9a} and Figure \ref{fig9b}. From the results, we can conclude that adding noise does not significantly affect the performance of the KST-GCNs and that the KST-GCNs are robust to the possible noise in the data.
\subsection{Model Interpretation}
In section \ref{Long-term}, we predict traffic speeds for the next 15 min, 30 min, 45 min, and 60 min based on historical data over a period of time using the KST-GCNs. To interpret the results in detail, we visualize the prediction results, which are shown in Figure \ref{fig10} to Figure \ref{fig13}. Note that, we only show the results of KF-T-GCN for the limited space. These results show the following:
\begin{figure}[htbp]
\centering
\includegraphics[width=0.85\linewidth]{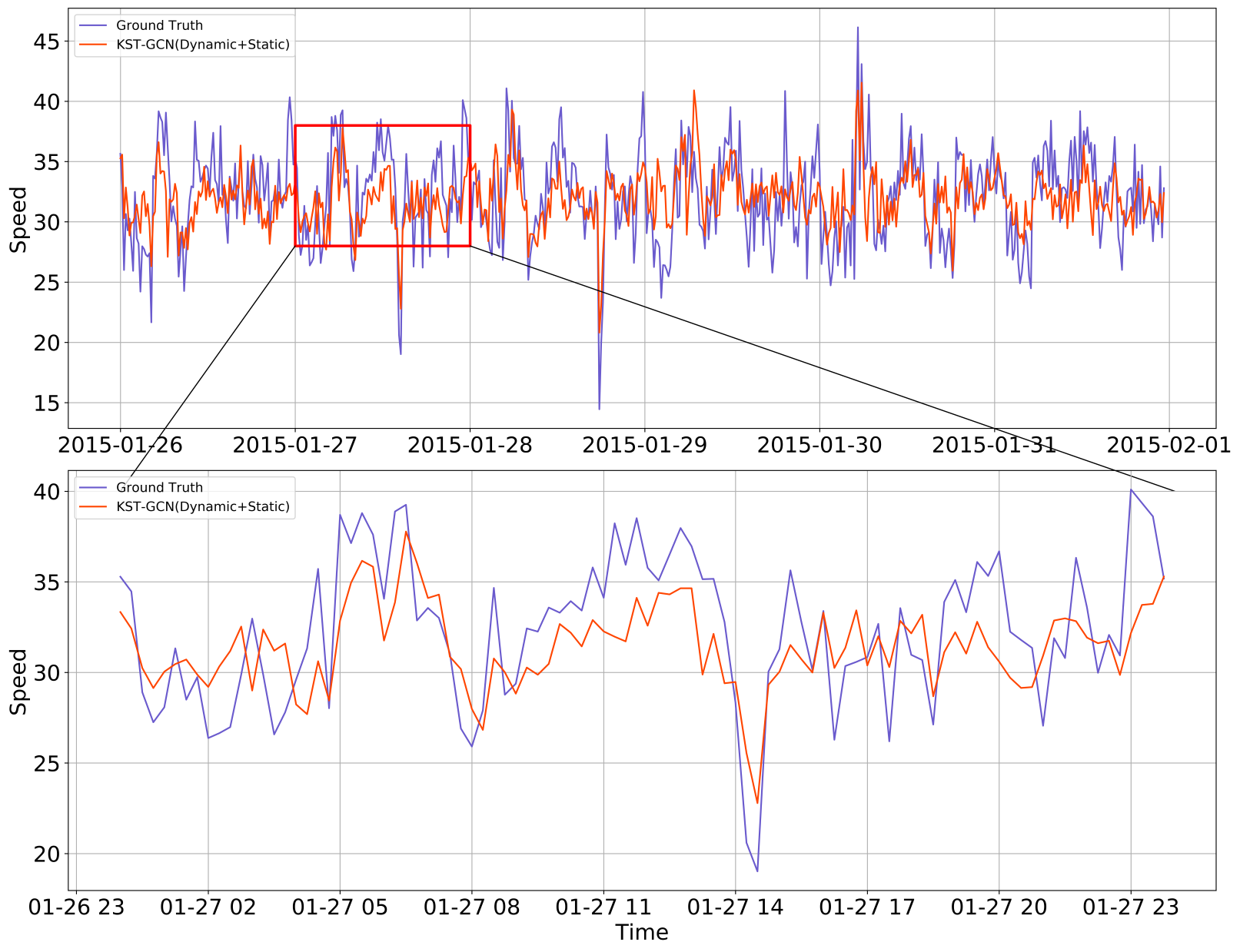} 
\caption{Result for prediction horizon of 15 minutes.}
\label{fig10}
\end{figure}
\begin{figure}[htbp]
\centering
\includegraphics[width=0.85\linewidth]{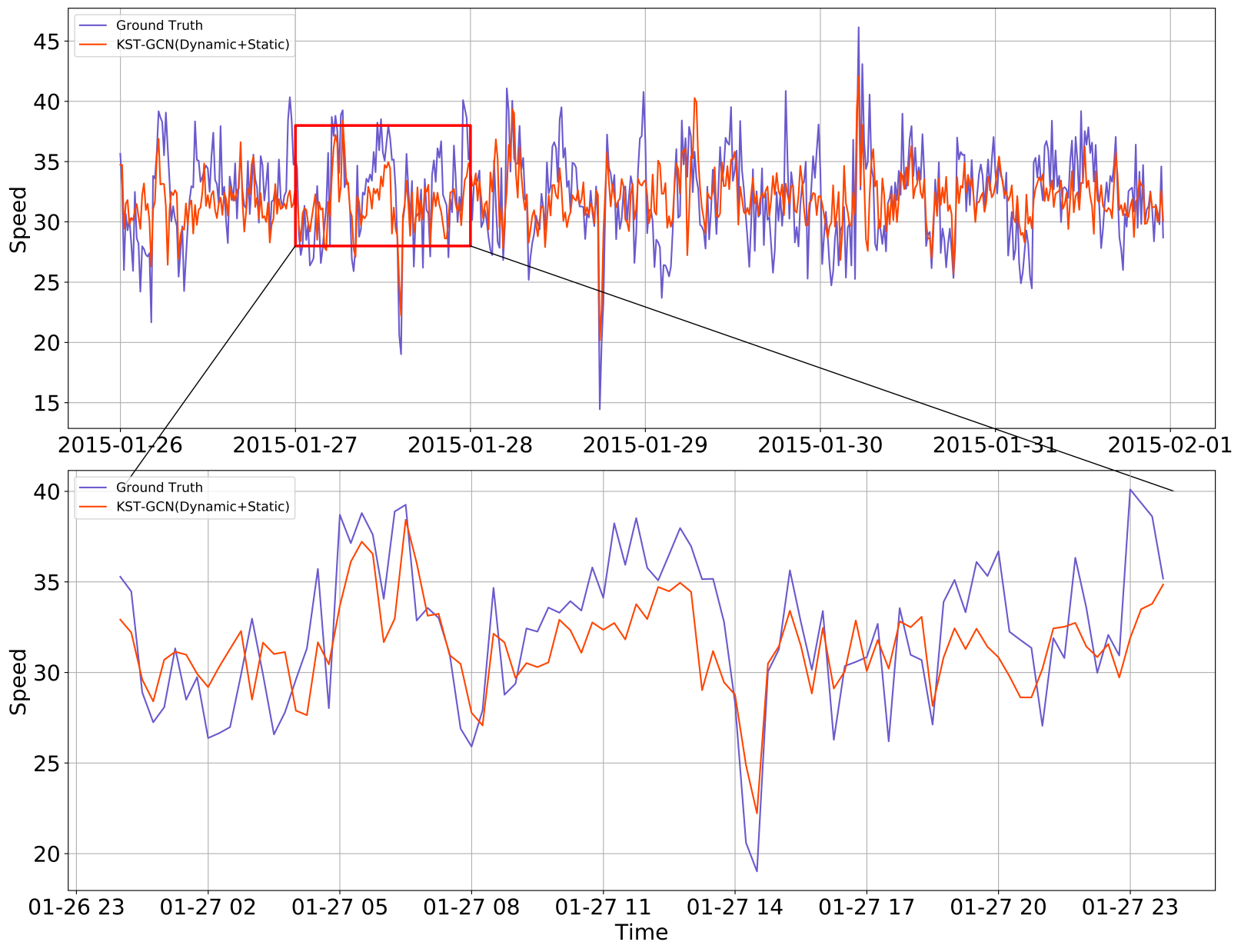} 
\caption{Result for prediction horizon of 30 minutes.}
\label{fig11}
\end{figure}
\begin{figure}[htbp]
\centering
\includegraphics[width=0.85\linewidth]{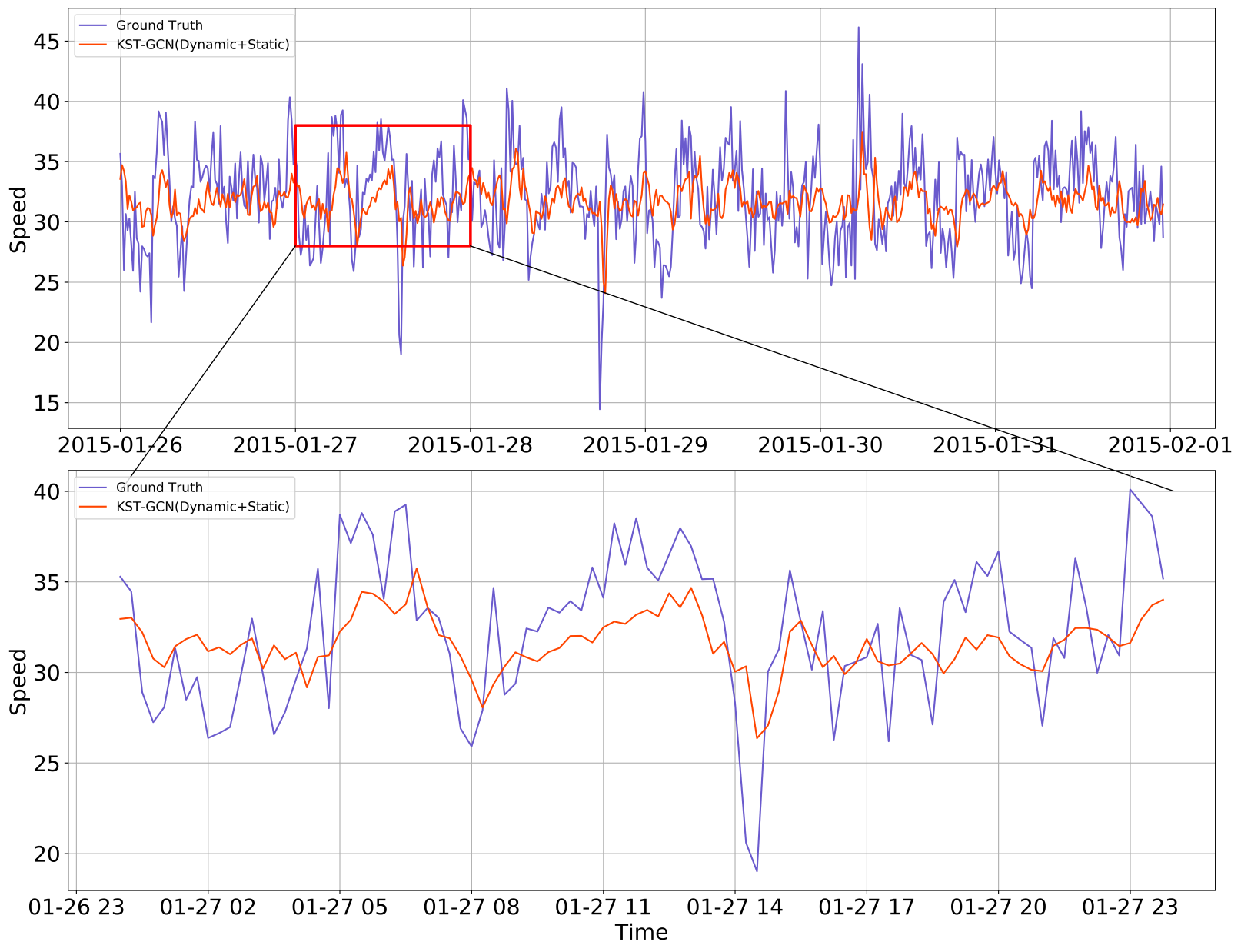} 
\caption{Result for prediction horizon of 45 minutes.}
\label{fig12}
\end{figure}
\begin{figure}[htbp]
\centering
\includegraphics[width=0.85\linewidth]{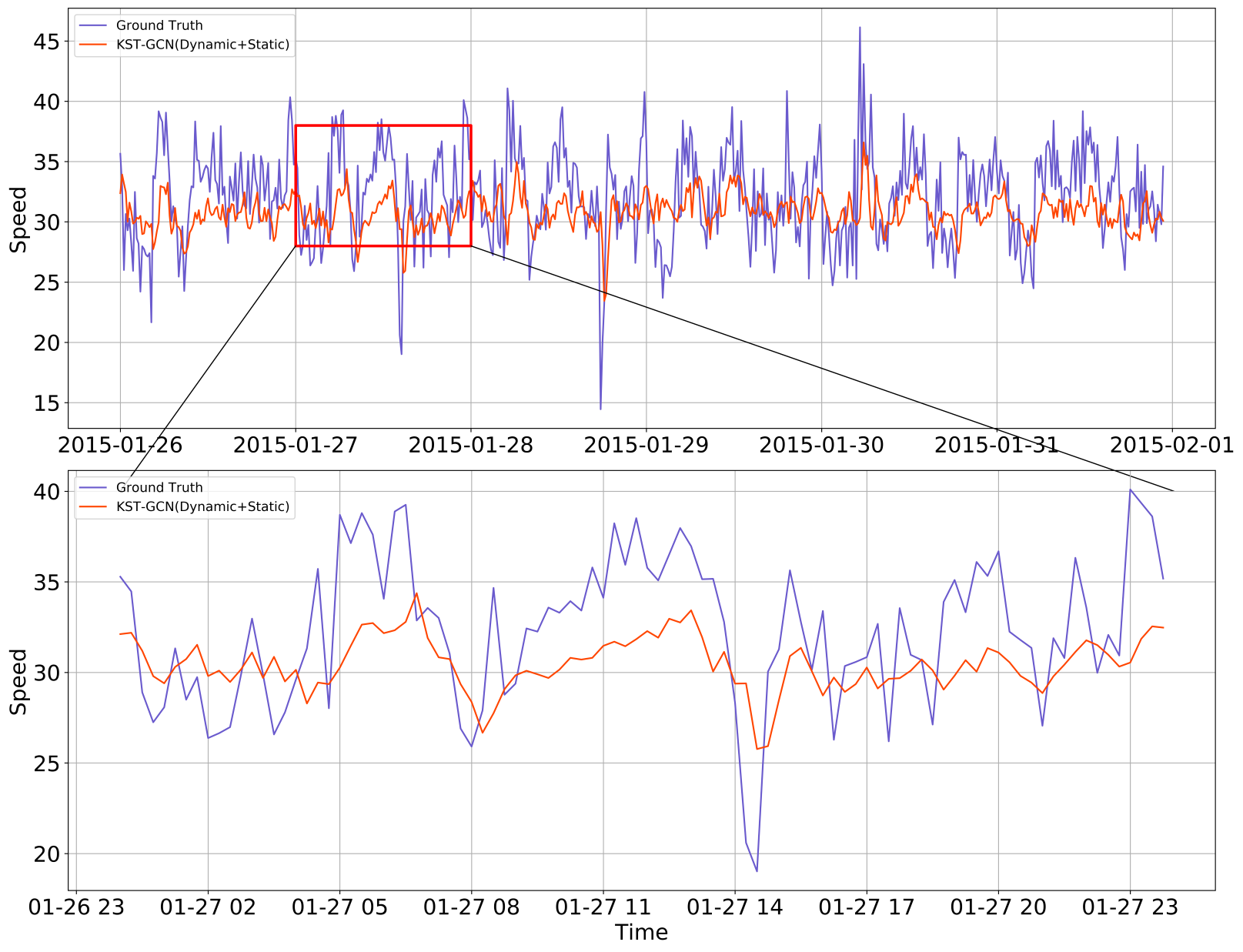} 
\caption{Result for prediction horizon of 60 minutes.}
\label{fig13}
\end{figure}

\begin{enumerate}
    \item For all prediction horizons, the KF-T-GCN well captures the change tendency of traffic data and is thus able to predict more accurately.
    \item The short-range prediction results of the KF-T-GCN are better than the long-range ones. The prediction result and the ground truth are closer in short-term prediction (15 min) than in long-term prediction (60 min), as shown in Figure \ref{fig10} and Figure \ref{fig13}. This may due to the more complex spatial-temporal dependencies in long-term prediction, and additional factors are needed to take into account.
    \item The prediction of the KF-T-GCN has significant deviations at turning points. The reason may be that the variations at peaks are influenced not only influenced by the weather and POIs but also by a combination of factors such as traffic contingencies.
\end{enumerate}

\section{Conclusion}
In this paper, we present the KST-GCNs, the knowledge-driven traffic forecasting models based on knowledge representations and backbone spatial-temporal graph convolutional networks. Our method tries to address the problem of neglecting correlations between traffic information and external factors in conventional urban traffic prediction methods. The KST-GCNs first adopt a knowledge graph representation method to distill the knowledge of correlations, then propose a Knowledge Fusion Cell (KF-Cell) to empower the backbone spatial-temporal graph convolutional network to perceive the knowledge. The experimental results demonstrate the superiority of KST-GCNs under various prediction horizons compared to their backbones. Additionally, the ablation and perturbation analysis verify the effectiveness and robustness of KST-GCNs. This study provides a feasible solution for incorporating multiple external factors in the traffic forecasting task. The knowledge graph can also facilitate traffic forecasting models other than KST-GCNs as long as the appropriate fusion mechanism is proposed. However, our method is a preliminary attempt that introduces knowledge graphs to exploit traffic semantic information and boost the forecasting performance, and there are many future works. Due to the difficulty of data collection, the knowledge graph constructed in this paper only considers POI and weather information and has limited expressive power, thus limiting the performance of KST-GCN. It is promising that KST-GCNs can be more effective when more multi-source data are available for constructing the knowledge graph.

\section*{Acknowledgments}
We appropriate anonymous reviewers for their valuable suggestions to make this paper better. This work was supported by the National Natural Science Foundation of China under Grant 42171458 and 61973047, and supported by Fundamental Research Funds for the Central Universities of Central South University under Grant 2019zzts881. We appropriate the High Performance Computing Platform of Central South University and HPC Central of Department of GIS to provide HPC resources.

\ifCLASSOPTIONcaptionsoff
  \newpage
\fi



\bibliographystyle{IEEEtran}
%



%
\bibliography{KST-GCN}




\end{document}